\pdfoutput=1

\documentclass[11pt]{article}

\usepackage{ACL2023}

\usepackage{times}
\usepackage{latexsym}

\usepackage[T1]{fontenc}

\usepackage[utf8]{inputenc}

\usepackage{microtype}

\usepackage{inconsolata}

\usepackage{hyperref}
\usepackage{url}
\usepackage{todonotes}
\usepackage{multirow}

\usepackage{amssymb}
\usepackage{xcolor}
\usepackage{url}
\usepackage{amsmath}
\usepackage{graphicx}
\usepackage{caption}
\usepackage{tabularx}

\title{Sign Language Sense Disambiguation}

\author{Jana Grimm\thanks{\;\; Equal contribution.} \and Miriam Winkler\footnotemark[1] \and Oliver Kraus\footnotemark[1] \and Tanalp Agustoslu\footnotemark[1]\\
  LMU Munich \\
  {\texttt{j.grimm@campus.lmu.de}} \and \texttt{mi.winkler@campus.lmu.de} \and \\ \texttt{o.kraus2@campus.lmu.de} \and \texttt{t.agustoslu@campus.lmu.de}\\ 
  \\}

\begin{document}
\maketitle
\begin{abstract}
This project explores methods to enhance sign language translation of German sign language, specifically focusing on disambiguation of homonyms. Sign language is ambiguous and understudied which is the basis for our experiments. We approach the improvement by training transformer-based models on various bodypart representations to shift the focus on said bodypart. To determine the impact of, e.g., the hand or mouth representations, we experiment with different combinations. The results show that focusing on the mouth increases the performance in small dataset settings while shifting the focus on the hands retrieves better results in larger dataset settings.
Our results contribute to better accessibility for non-hearing persons by improving the systems powering digital assistants, enabling a more accurate interaction. The code for this project can be found on \href{https://github.com/OvrK12/slt}{GitHub}.

\end{abstract}

\section{Introduction}
People communicate not only through spoken language but also through facial expressions and gestures. This form of expression is rooted in the early days of humanity when people did not yet communicate with spoken language as we know it today 
\cite{tomasello2009Ursprünge}.
Especially for non-hearing people, the face, hand, and body movements are substantial parts of communication and the meaning of the signs is built through different combinations of those components \cite{Braem2001The_Hands}.
However, just as in spoken language, the meaning of hand signs is not always clear from the start and the meaning might only become clear by looking at the other body and face movements.
This encourages our experiments to dive deeper into sign language translation, especially since improving the performance of such models and advancing the research aids the accessibility for deaf people in different domains. For example, this could enable non-hearing people to interact with digital assistants via video by signing.

The specific underlying question we want to explore is: \textit{Can we improve the translation of ambiguous signs by manually setting the focus on different body parts in sign language videos?} To examine the influence of our experiments in regard to disambiguation, we want to focus our analyses on homonyms. 
Homonyms are signs that have the same body but can have different meanings. For example, the German sign for "food" ("Essen") has multiple meanings in the domain of food like "breakfast" ("Frühstück"), "lunch" ("Mittagessen"), or "dinner" ("Abendbrot"). 
Another meaning of the sign for "food" mirrors the written word for "food" ("Essen") in German which can also point to the German city Essen.\footnote{Source: \href{https://www.sign-lang.uni-hamburg.de/korpusdict/bags/bag80.html\#reading2open}{DW-DGS}} Sign language speakers disambiguate these shortcomings by mouthing the word of the spoken language \cite{Quer2015Ambiguities}.
This leads to the assumption that ambiguities in sign language could be dissolved by incorporating information about the mouth in the input of a model.\\
Our experiments incorporate training a transformer-based model \cite{Camgöz2020Sign} on representations of different parts of sign videos, e.g. the hands and the mouth. In various combinations of training setups, we explore the impact of different body parts on disambiguation.


\section{Related Work}
\subsection{Sign Language Tasks}
Working with multimodal sign language data entails three important tasks: sign segmentation, recognition, and translation \cite{Camgöz2020Sign}.
The segmentation task detects the borders of sign language videos and extracts the individual signs. Even though this is the base for the following tasks, sign segmentation is severely understudied.
Sign language recognition assigns matching glosses to the detected signs. A gloss is a linguistic transcription of the signs but does not form a fluent textual translation of the spoken sentence. Examples of glosses can be seen in Figure \ref{fig:matching}. This is the hardest task from a computer vision perspective because signs are highly dependent on their dimensionality, which is not completely conveyed in a two-dimensional video. The glosses are then embedded together as a sentence representation.
This sentence representation is the input for the sign language translation task which processes the information and transforms it into a spoken language sentence, similar to text-based machine translation. It can happen in two ways: You can produce a video of signs from a text \cite{Stoll2018Sign} or extract a written text as a transcription of a sign language video \cite{Cui2023Spatial, Camgöz2020Sign}. This project focuses on extracting text from sign language image sequences. \\
The two relevant tasks for our project are sign language recognition and sign language translation. While both tasks share similarities, it is important to distinguish them. The transformer employed in our project performs recognition and translation jointly, but we focus on improving only the translation task. More details on this setup can be found in Section \ref{sec:methods}.

\subsection{Sign Language Transformers}
Our experimental setup is based on the work of \citet{Camgöz2020Sign}. They train a transformer-based model to perform joint sign language recognition and translation. This improves the performance when compared to a separate encoder structure.
In contrast, \citet{Cui2023Spatial} work with a transformer-based continuous sign language recognition (CSLR) method where they model global and spatial action features and semantic features separately. The similarity between the sign image sequences and the generated text is calculated via cross-entropy loss. \\
Both approaches show that employing a transformer-based model brings improvements compared to the state-of-the-art experiments that use convolutional neural network-based architectures \cite{Cheng2020Fully}. \\ 
We want to extend these improvements with our experimental setup which is further explained in Section \ref{sec:experiment}.

\subsection{Sign Language Resources}
This project uses a corpus of German sign language. Sign language is a complex language of its own that is spoken by about 80,000 speakers \cite{vogel2024deutscher}. The research in Germany only started in the 1970s in Hamburg and in 1987, the first institution for deaf communication was founded \cite{jacobi2020barrierefreie}. Since then, sign language research gradually started to increase and especially the University of Hamburg is well known for its renowned research. It is also them who created a big online sign language resource, the `Meine DGS'\footnote{\href{https://www.sign-lang.uni-hamburg.de/korpusdict/overview/index-dgs.html}{Meine DGS}} online dictionary \cite{schulder2024statement}. It contains videos, glosses, and many other types of annotation. We use this data for reference and to find homonyms for our project evaluation.
The real-life data from `Meine DGS' proves valuable to us as our data - RWTH-PHOENIX-Weather Database of German Sign Language \cite{Koller2015Continuous} - also stems from real-life data, namely the German television channel PHOENIX. The videos were labeled manually by native speakers. The benefit of using this data is the applicability to real-life signing, as most available sign language corpora are recorded primarily for research purposes which leads to a different kind of signing than one would encounter in actual sign language communication.



\section{Methods}
\label{sec:methods}
\subsection{Dataset}
We use the RWTH-PHOENIX-Weather Database of German Sign Language \cite{Koller2015Continuous} for training and evaluation of the transformer. This dataset contains videos of sign language speakers translating the weather forecast shown by the German TV channel PHOENIX and their translations. For 266 of the included sentences, the position of the face and the hands have been annotated. We use this portion of the data to produce a dataset fitted to test the hypothesis that incorporating more information about different body parts improves the disambiguation of homonyms. We focus on the videos which have annotations for hand and face tracking to correctly extract the hands and face from the videos. This procedure ensures that the input for our experiments is based on gold standard and prevents incorrectly extracted body parts.

\subsection{Data Preprocessinng}
\subsubsection{Baseline Data Preparation}
We preprocess the 266 sentences in such a way that it is formatted as a suitable input for the transformer model. We extract the following information from the available metadata: video name, video representation, signer, gloss and text (spoken language sentence). The video representation is a tensor of the pillow \cite{clark2015pillow} representations of the singular frames. This baseline dataset is split into train, test and dev splits following the ratio of 70-20-10.

\begin{figure}
    \centering
    \includegraphics[width=\linewidth]{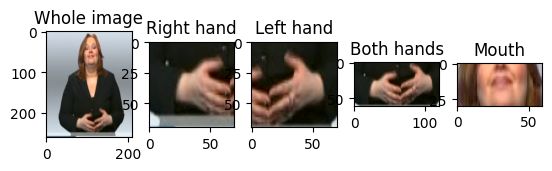}
    \caption{Example for body part extraction.}
    \label{fig:bodypart_pics}
\end{figure}

\subsubsection{Body Part Extraction}
We want to enhance the recognition of ambiguous sentences by manually shifting the transformer's focus on different body parts.
We do this by additionally feeding the tensors of the extracted body parts into the transformer at training time. \\
We extract the body parts by the coordinates provided by the dataset \cite{Koller2015Continuous} to locate the positions of the right hand, left hand and the nose. We extract them with the pillow Python package \cite{clark2015pillow}. An example image and the cropped frames can be seen in Figure \ref{fig:bodypart_pics}. \\
With the cut-outs of the bodyparts, we then go through the same procedure as for our baseline data and create datasets in addition to our baseline data for the following combinations:
\begin{itemize}
    \item Mouth and whole picture
    \item Both hands and whole picture
    \item Mouth, both hands and whole picture
\end{itemize}
The bodypart data thus contains the same meta data as the baseline data, except for the image representations. To create a joint representation of the information coming from multiple pictures, we multiply the vectors. This keeps the dimensionality the same, but adds the additional information. \\

\subsection{Signjoey - Sign Language Transformer}
\label{sec:experiment}
Our transformer setup is based on the work by \citet{Camgöz2020Sign}, which introduced a unified transformer architecture for both sign language recognition and translation tasks. This architecture builds upon the Joey NMT framework \citep{kreutzer-etal-2019-joey}, leveraging the Connectionist Temporal Classification (CTC) loss function to enable joint learning and training in an end-to-end manner. The CTC loss function addresses the challenge of ground-truth timing information, often missing in unsegmented time-series data such as videos or audio sequences. In our context, this means we do not have precise information about when each sign begins and ends within a video. Accurate timing information is crucial for correctly labeling sequences in time-series data, which is essential for training sign language recognition models that rely on supervised learning.










We explore different hyperparameter spaces as listed in Table \ref{tab:Hyperparameters}. As we find that the hyperparameter settings have a big influence on the results, we change them accordingly to the reported experiment.

\begin{table}[h!]
    \centering
    \footnotesize
    \begin{tabular}{|c|c c c|} \hline 
         \textbf{Params} & \textbf{Values} & & \\ \hline 
         Epochs & 5 & 10 & 20 \\ \hline
         Hidden Dims & 256 & 512 & \\ \hline
         Batch Size & 16 & 32 & 64\\ \hline 
         Learning Rate (LR) & \(1 \times 10^{-3}\) & \(1 \times 10^{-4}\) & \(1 \times 10^{-5}\) \\ \hline 
    \end{tabular}
    \caption{Explored hyperparameter space}
    \label{tab:Hyperparameters}
\end{table}


\section{Evaluation Results}
To conduct an analysis on the disambiguation capabilities of a transformer-based model when presented with information about different body parts like the mouth and the hands, we explore different dataset and model variations to find the best foundation (see Section \ref{sec:data_model_experiments}). We then use this foundation to examine in which way the models performance is influenced by different body parts (Section \ref{sec:body_results}) and conclude by investigating the homonyms in our datasets and the effect on their translation in Section \ref{sec:homonym_analysis}.  

\subsection{Data and Model Variations}
\label{sec:data_model_experiments}
\begin{figure}
    \centering
    \includegraphics[width=\linewidth]{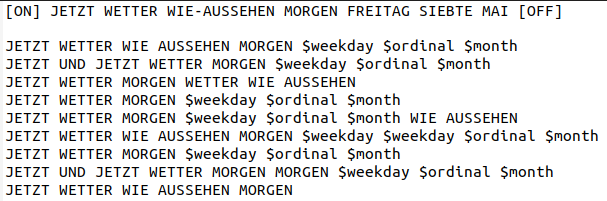}
    \caption{One example of the matching problem in the RWTH-PHOENIX-Weather Database of German Sign Language \citep{Koller2015Continuous}. The gloss on the top is the original gloss aligned with the video, the glosses on the bottom are possible matching glosses aligned with the textual translation of the video.}
    \label{fig:matching}
\end{figure}

When training the transformers on our data, there are two challenges: the small number of instances in the dataset (only 266 videos for training, development, and testing) and the construction of the RWTH-PHOENIX-Weather Database of German Sign Language \citep{Koller2015Continuous} itself. The structure of the RWTH-PHOENIX-Weather Database \citep{Koller2015Continuous} is not a single file but split into different parts which can be used for various sign language modeling tasks. When using it for sign language translation, you need to match the glosses in two different files that have been processed without any documentation on how they have been processed.\footnote{The authors state on their website \citep{forster-2012-rwth-website} that they provide the unprocessed data when enquired but unfortunately we could not reach them.} Figure \ref{fig:matching} displays an example of the original gloss and processed glosses, that could match the original gloss. To match the original and the processed gloss, we split them up into wordsets and retrieve the textual translation for the gloss with the highest number of intersected words between the two sets. As a more complex method, we further experiment with computing the BLEU score between two glosses instead of matching them via intersected word sets. Due to the anonymization during the processing step of the glosses, we find that deanonymizing the textual translations with GPT-4.o \citep{openai2024gpt4technicalreport} improves the predictions of the models by approximately 2-4 BLEU points for BLEU-1 and BLEU-2 (see Table \ref{tab:match_experiments}). We further experiment with using a GPT-4.o \citep{openai2024gpt4technicalreport} to create textual translations from the unprocessed gloss to bypass the matching probability. To increase the number of instances, we augment the already created datasets by flipping the images and duplicating them with a new contrast rate. The augmented data set contains 798 instances with a 70-20-10 split between training, testing, and development set.\\
The results in Table \ref{tab:match_experiments} show that the augmentation of the dataset lowers the perplexity of the model (PPL). The translations on the character level benefit from a bigger augmented dataset (CHRF). On the word level deanonymizing proves to be useful (BLEU-1, BLEU-2, ROUGE). The translations generated by GPT-4.o \citep{openai2024gpt4technicalreport} without any matching perform worst, which leads to the assumption that human data is more useful than synthetic data in this case. According to the good performance of the BLEU-alignment and the wordset matching with GPT substitutions, we base our following experiments on them.\\\\
To eradicate the missing image processing capability of the transformer due to the small training data, we utilize two different methods. Instead of presenting the images as PIL-images \citep{clark2015pillow} to the model as done in the original setup (see Section \ref{sec:data_model_experiments}), we preprocess them with EfficientNet \citep{tan2020efficientnetrethinkingmodelscaling}\footnote{We use wordset+GPT without data augmentation as foundation for our experiments with EfficientNet \citep{tan2020efficientnetrethinkingmodelscaling} because the augmentation library \href{https://albumentations.ai/}{albumentations} removes already preprocessed information again for normalization purposes.} to enhance the models understanding of the images. The second method is to pre-train the transformer with the RWTH-PHOENIX-Weather 2014T \citep{Camgoz_2018_CVPR} and fine-tune it with our dataset. 
The data scarcity also impacts this part of our experiments. As there is no fine-tuning script available, we write the script considering the size mismatch issue that is caused by the unmatched size of pre-training data and fine-tuning data that results in different parameter sizes (the pre-trained model has 30M parameters, whereas during fine-tuning we only make use of 23M parameters.) To counteract this issue we follow two approaches. One of them is to combine all data we have and continue with the training. As our own data is underrepresented within the whole combined data in this training setting, we observe decrease in translation quality. On the other hand, as a second approach, we first pre-train the model and then fine-tune using our own augmented and processed data. To execute this in a more safe manner, we first try to freeze certain layers and gradually unfreeze them during the training, but as this does not solve the size mismatch problem completely, we instead partially reinitialize some of the layers using uniform distribution and expect from the model to use the previous weights to adapt itself to the new data.\\
The results in Table \ref{tab:fine_experiments} show the capability of fine-tuning clearly. Compared to the results in Table \ref{tab:match_experiments} preprocessing the images retrieves comparable results. In contrast to fine-tuning on the other hand, we observe gaps of approx. 53-65 BLEU. With our experiments for wordset matching and EfficientNet \citep{tan2020efficientnetrethinkingmodelscaling} we could not retrieve any BLEU > 0 for more than bigrams (except for BLEU-align and BLEU-3, see Table \ref{sec:body_results}), fine-tuning on the other hand, achieves comparable scores for 3- and 4-gram matches (see Table \ref{sec:body_results}). It also beats all of the previous results for ROUGE and CHRF. Only the perplexity is higher than wordset + GPT + aug which could stem from the differing dataset used for pre-training and the validation. Following these results, we focus on fine-tuning to explore if the extraction of the mouth can be used as a disambiguation feature for sign language. We also include BLEU-align + GPT + aug to make the influence of fine-tuning on the incorporation of body parts into the data recognizable.\\

\begin{table*}
    \centering
    \begin{tabular}{|l|c c c c c|} \hline 
         \textbf{Experiment} & \textbf{BLEU-1} & \textbf{BLEU-2} & \textbf{ROUGE} & \textbf{CHRF} & \textbf{PPL}\\ \hline  
         wordset + GPT & 14.67 & 4.08 & \textbf{16.39} & 13.10 & 215.28 \\
         wordset + GPT + aug & 12.74 & 2.90 & 13.66 & \textbf{22.93} & \textbf{3.16} \\ 
         BLEU-align + GPT + aug & \textbf{16.00} & \textbf{5.90} & 14.79 & 19.09 & 191.82 \\ 
         wordset & 12.50 & 0 & 13.91 & 3.93 & 157.41 \\ 
         GPT (trans) + aug & 4.31 & 0 & 5.65 & 2.63 & 134.57 \\ \hline 
    \end{tabular}
    \caption{Results of the matching experiments on the baseline dataset tested on the held out instances of the RWTH-PHOENIX-Weather Database of German Sign Language \citep{Koller2015Continuous} trained with 5 epochs, 512 hidden dimensions, a batch size of 32 and a learning rate of \(1 \times 10^{-3}\). (wordset: matching via intersections, GPT: deanonymizing, aug: augmentation, BLEU-align: matching via BLEU-alignment, GPT (trans): textual translations of the glosses without gloss matching}
    \label{tab:match_experiments}
\end{table*}

\begin{table*}
    \centering
    \footnotesize
    \begin{tabular}{|l|c c c c c c|} \hline 
         \textbf{Experiment} & \textbf{BLEU-1} & \textbf{BLEU-2} & \textbf{BLEU-3} & \textbf{ROUGE} & \textbf{CHRF} & \textbf{PPL}\\ \hline  
         EfNet + wordset + GPT & 15.27 & 3.75 & 0 & 15.67 & 11.23 & 165.03 \\
         fine-tune + BLEU-align + GPT + aug & \textbf{68.60} & \textbf{66.37} & \textbf{65.48} & \textbf{70.60} & \textbf{71.98} & \textbf{5.17}\\ \hline 
    \end{tabular}
    \caption{Results of the baseline dataset for fine-tuning and efficient image processing tested on the held out instances of the RWTH-PHOENIX-Weather Database of German Sign Language \citep{Koller2015Continuous}, trained with 5 epochs, 512 hidden dimensions, a batch size of 32 and a learning rate of \(1 \times 10^{-3}\) for EfNet + wordset + GPT and 40 epochs, 512 hidden dimensions, a batch size of 64 and a learning rate of \(1 \times 10^{-3}\) for fine-tune + BLEU-align + GPT + aug. (EfNet: EfficientNet, wordset: matching via intersections, GPT: deanonymizing, aug: augmentation, fine-tune: fine-tuning, BLEU-align: matching via BLEU-alignment)}
    \label{tab:fine_experiments}
\end{table*}

\subsection{Integration of body part representations}
\label{sec:body_results}
We report the results of the different body part experiments on two model setups: BLEU-alignment with and without fine-tuning. As pointed out in Section \ref{sec:data_model_experiments}, the alignment of the different parts in the dataset via BLEU combined with GPT-4.o \citep{openai2024gpt4technicalreport} serves as a good foundation for sign language translation with the slt-Transformer \citep{Camgöz2020Sign} even on small datasets. We compare the results with and without fine-tuning to test if the information about the different body parts gets lost in the information conducted from the pre-training without body parts.\\

\begin{table*}[!ht]
    \centering
    \footnotesize
    \begin{tabular}{|l|ccccccc|} \hline 
         \textbf{BLEU-align + GPT + aug} & \textbf{BLEU-1} & \textbf{BLEU-2} & \textbf{BLEU-3} & \textbf{BLEU-4} & \textbf{ROUGE} & \textbf{CHRF}\\ \hline  
         Baseline & 10.28 & 5.66 & 4.12 & \textbf{3.31} & 10.91 & 17.82\\
         Mouth & \textbf{20.65} & \textbf{8.37} & \textbf{4.40}  & 1.93 & \textbf{20.01} & \textbf{19.14}\\
         Hands & 15.39 & 5.18 & 1.90 & 0 & 15.65 & 17.29\\ 
         Hands + Mouth & 13.87 & 3.79 & 1.06 & 0 & 16.31 & 12.48\\ \hline 
         \textbf{fine-tune + BLEU-align + GPT + aug} & \textbf{BLEU-1} & \textbf{BLEU-2} & \textbf{BLEU-3} & \textbf{BLEU-4} & \textbf{ROUGE} & \textbf{CHRF}\\ \hline  
         Baseline & 68.60 & 66.37 & 65.48 & 64.93 & 70.60 & 71.98\\
         Mouth & 68.65 & 66.79 & 66.12 & 65.80 & 68.58 & 72.24\\
         Hands & \textbf{76.38} & \textbf{74.85} & \textbf{74.20} & \textbf{73.81} & \textbf{77.25} & \textbf{80.10}\\ 
         Hands + Mouth & 73.27 & 71.14 & 70.25 & 69.76 & 74.06 & 75.73\\ \hline 
    \end{tabular}
    \caption{Performance on the model BLEU-align + GPT + aug with information about different body parts with and without fine-tuning tested on the held out instances of the RWTH-PHOENIX-Weather Database of German Sign Language \citep{Koller2015Continuous} trained with 20 epochs, 512 hidden dimensions, a batch size of 32 and a learning rate of \(1 \times 10^{-4}\) for BLEU-align + GPT + aug and 40 epochs, 512 hidden dimensions, a batch size of 64 and a learning rate of \(1 \times 10^{-3}\) for fine-tune + BLEU-align + GPT + aug. (GPT: deanonymizing, aug: augmentation, fine-tune: fine-tuning, BLEU-align: matching via BLEU-alignment)}
    \label{tab:body_results}
\end{table*}

Table \ref{tab:body_results} shows differing results for a small and a large training setup. For the small setup BLEU-align + GPT + aug, we can see that adding information about different body parts improves the results for Mouth, Hands, and Hands + Mouth, while Mouth performs best on word level with differing n-gram sizes and on character level. As seen in our previous experiments (Section \ref{sec:data_model_experiments}), fine-tuning outperforms the models, that are trained only on our dataset, by far. Here, we can see, that the model tends to retrieve better results when body parts, especially the hands are included. This leads to the assumption that the model can learn the most information from the mouth when only a small amount of training data is provided. However, when the model can process more data, focusing on the hands yields the best results. This is likely because hands are the most crucial body part for conveying information in sign language. Consequently, incorporating hand data leads to the most significant improvement in the model's performance. When comparing the Mouth Transformer with the Baseline, one can see, that incorporating data about the mouth increases the score only marginally. This small change could come from the small percentage of body part data, that the model is trained with, in comparison to the size of the pre-training data. The model might not have been able to draw a lot of new information from the fine-tuning. We also need to take into account that the hyperparameter configuration has a big influence on the outcome. Therefore, it would make sense to compare the results when each body part experiment is tuned for its respective best hyperparameter setting.\\
The results display that the hypothesis of incorporating additional information about body parts into the task of sign language translation improves the overall performance. They further show that the outcome depends highly on the setting. As computational and time-limiting factors are often important variables when conducting research, our comparison reveals how to deal with small data settings. 

\begin{figure}
  \centering
    \includegraphics[width=\linewidth]{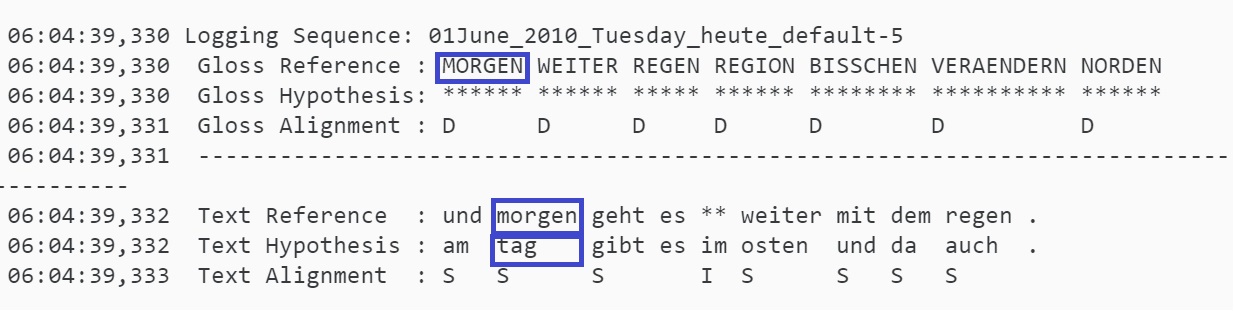}
    \caption{One example of the translation of our best-performing model. The ambiguous sign "tomorrow" ("morgen") is translated as "day" ("tag") incorrectly.}
    \label{fig:translation}
\end{figure}

\subsection{Homonym analysis}
We extract homonyms from the 'Meine DGS' webpage and match them with the glosses in our dataset. Our analysis reveals that 170 out of 266 glosses contain homonyms. After manually adding the homonyms that our script initially missed, we find an average of 1.12 homonyms per gloss. We discard irrelevant homonyms that are not found in our dataset such as "Sunday" ("Sonntag") and "church" ("Kirche") and only select 6 of them. However, after carefully processing data and finding the best working hyperparameters, we still observe in our experiments that even our best-performing model struggles to capture nuanced details such as homonyms (see Figure \ref{fig:translation}).

As mentioned previously, due to computational limitations and data constraints, achieving better results is currently challenging. However, we believe that by integrating the methods outlined in our future work, the underexplored area of homonyms in this domain can be further examined, potentially leading to improved outcomes.

\label{sec:homonym_analysis}

\section{Limitations and Future Work}
There are different challenges and limitations to our experiment. The computational power and restricted time due to project deadlines force us to train the models on a very small dataset. Even though we fight this issue back by fine-tuning and augmentation, the robustness of the experiment is not given when trained on such a small amount of data. Therefore, it would make sense to rerun our experiment on a bigger dataset. Despite our efforts to address data limitations as thoroughly as possible, there is still room for future work. 
One suggestion by \citet{Wong2024Sign2GPT} is to integrate pseudo-glosses using additional data augmentation methods, such as random resized cropping and color jitter, with image representations.\\
Furthermore, we do not have access to the top-performing image representation from the original paper (Inception v4 in a CNN+LSTM+HMM setup) \cite{Camgöz2020Sign}, which achieved a 17 percent improvement in WER over the best-performing EfficientNet B7, as it is not open-source. Incorporating this model could also potentially yield better results.\\
The missing link between the video and the textual translation faces the model with the challenge of possibly learning incorrect translations. Even though, BLEU-alignment with GPT4.o \citep{openai2024gpt4technicalreport} substitutions can partially cope with this problem, the model still is limited to a error-prone dataset. We therefore plead to create datasets with body part annotation and direct links to the textual translations.\\
Another problem regarding the dataset is that it only covers weather forecasts which are quite repetitive. This is a challenge for the model because it tends to overfit quickly. It would be interesting to repeat the experiment on different domains to explore the generalization capabilities.\\\\
Furthermore, there are different types of ambiguities in sign language and ways for sign language users to solve them such as contextual dependencies and scope ambiguity. In this experiment, we only focus on one aspect of ambiguity (homonyms) and one way to solve them (addition of mouthing). For future work, it would be interesting to investigate different types of (dis)ambiguations and to incorporate the input of sign language users on this topic.


\section{Conclusions}
While textual Machine Translation is a well-researched NLP task, we focus in this project on a more complex multimodal part of this topic: Sign Language Translation. We conduct experiments to improve the disambiguation capabilities of the state-of-the-art sign language transformer introduced by \citet{Camgöz2020Sign}. The basis of our experiments mirrors homonym disambiguation techniques of sign language speakers: the mouthing of ambiguous words. While the incorporation of information about the mouth proves to increase the performance of the model, we find information about the hands to be more useful on a large scale. In our homonym analysis, we find out that even though the translation quality is improved by incorporating hands and mouth information, there is still room for improvement, as our best-performing model currently cannot distinguish between homonyms. 
For future projects, we propose to repeat the experiment with a bigger dataset that integrates multiple domains, to test whether incorporating body part information increases the generalization and is scalable.

\bibliography{main}
\bibliographystyle{acl_natbib}

\appendix



\end{document}